\documentclass[conference]{IEEEtran}
\IEEEoverridecommandlockouts

\usepackage{cite}
\usepackage{amsmath,amssymb,amsfonts}
\usepackage{algorithmic}
\usepackage{graphicx}
\usepackage{textcomp}
\usepackage{xcolor}
\usepackage{color}
\usepackage{multirow}
\usepackage{booktabs}
\usepackage{colortbl}
\usepackage{url}
\usepackage{diagbox}

\def\BibTeX{{\rm B\kern-.05em{\sc i\kern-.025em b}\kern-.08em
    T\kern-.1667em\lower.7ex\hbox{E}\kern-.125emX}}
\begin{document}

\title{RemoteTrimmer: Adaptive Structural Pruning for Remote Sensing Image Classification
\thanks{Guangwenjie Zou and Liang Yao contributed equally to this work. Corresponding author: Fan Liu.}
\thanks{This work was partially supported by the Fundamental Research Funds for the Central Universities (No. B240201077), National Nature Science Foundation of China (No. 62372155 and No. 62302149), Aeronautical Science Fund (No. 2022Z071108001), Joint Fund of Ministry of Education for Equipment Pre-research (No. 8091B022123), Water Science and Technology Project of Jiangsu Province under grant No. 2021063, Qinglan Project of Jiangsu Province, Changzhou science and technology project No. 20231313. The work of Liang Yao was supported in part by Postgraduate Research \& Practice Innovation Program of Jiangsu Province (No. SJCX24\_0183).}
\thanks{Our code is available at: https://github.com/1e12Leon/RemoteTrimmer.}
}

\author{\IEEEauthorblockN{1\textsuperscript{st} Guangwenjie Zou}
\IEEEauthorblockA{\textit{Hohai University} \\
Nanjing, China \\
guangwenjiezou@hhu.edu.cn} \\
\IEEEauthorblockN{5\textsuperscript{th} Xin Li}
\IEEEauthorblockA{\textit{Hohai University} \\
Nanjing, China \\
li-xin@hhu.edu.cn} \\
\and
\IEEEauthorblockN{1\textsuperscript{st} Liang Yao}
\IEEEauthorblockA{\textit{Hohai University} \\
Nanjing, China \\
liangyao@hhu.edu.cn} \\
\IEEEauthorblockN{6\textsuperscript{th} Ning Chen}
\IEEEauthorblockA{\textit{Hohai University} \\
Nanjing, China \\
cn@hhu.edu.cn}\\
\and
\IEEEauthorblockN{3\textsuperscript{rd} Fan Liu}
\IEEEauthorblockA{\textit{Hohai University} \\
Nanjing, China \\
fanliu@hhu.edu.cn} \\
\IEEEauthorblockN{7\textsuperscript{th} Shengxiang Xu}
\IEEEauthorblockA{\textit{Hohai University} \\
Nanjing, China \\
xushx@hhu.edu.cn}\\
\and
\IEEEauthorblockN{4\textsuperscript{th} Chuanyi Zhang}
\IEEEauthorblockA{\textit{Hohai University} \\
Nanjing, China \\
zhangchuanyi@hhu.edu.cn} \\
\IEEEauthorblockN{8\textsuperscript{th} Jun Zhou}
\IEEEauthorblockA{\textit{Griffith University} \\
South East Queensland, Australia \\
jun.zhou@griffith.edu.au} \\
}

\maketitle

\begin{abstract}
Since high resolution remote sensing image classification often requires a relatively high computation complexity, lightweight models tend to be practical and efficient.
Model pruning is an effective method for model compression. However, existing methods rarely take into account the specificity of remote sensing images, resulting in significant accuracy loss after pruning. To this end, we propose an effective structural pruning approach for remote sensing image classification. Specifically, a pruning strategy that amplifies the differences in channel importance of the model is introduced. Then an adaptive mining loss function is designed for the fine-tuning process of the pruned model. Finally, we conducted experiments on two remote sensing classification datasets. The experimental results demonstrate that our method achieves minimal accuracy loss after compressing remote sensing classification models, achieving state-of-the-art (SoTA) performance. 
\end{abstract}


\section{Introduction}
Remote sensing technology has attracted significant attention due to its ability to remotely capture diverse and extensive environmental data~\cite{yan2015national, westberry2023gross, wang2023remote, han2023survey, an2024geological}. 
As one of the most significant techniques in remote sensing, image classification has been applied to agricultural crop monitoring, urban planning, and biodiversity assessment~\cite{zhu2019high, mehmood2022remote, zhang2022review, ma2024transfer}.
Despite its extensive applicability, Remote Sensing Image Classification (RSIC) still faces several challenges.
For example, remote sensing images typically have high resolution. To ensure classification accuracy, mainstream approaches classify images by dividing them into smaller patches. However, this divide-and-conquer strategy exponentially increases inference time.
Meanwhile, high-precision classification models typically involve numerous parameters, further escalating the model inference time.
These challenges limit the efficiency and scalability of RSIC.

\begin{figure}[t]
\centerline{\includegraphics[width=1\linewidth]{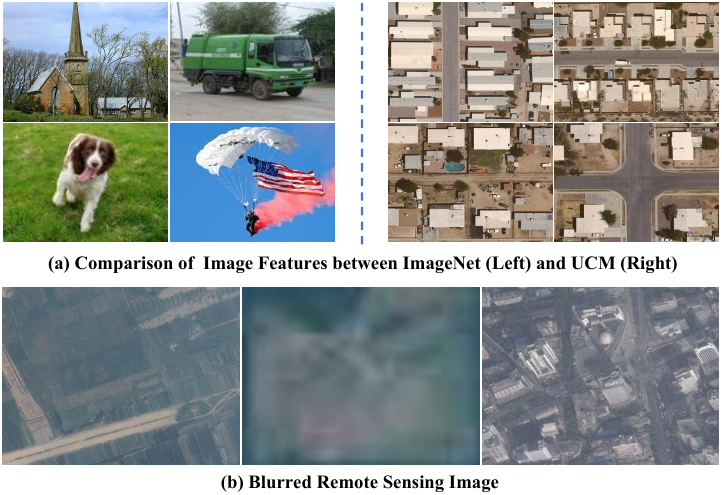}}
\caption{Motivations of our proposed method. \textbf{(a)} Comparison of general (ImageNet) and remote sensing (UCM) images. Remote sensing images have top-down views from different heights, resulting in greater scale variations of the objects. This situation usually leads to the narrowing of differences between similar object features.
\textbf{(b)} Remote sensing images often suffer from unclear images with atmospheric noise pollution, which makes lightweight models learning more difficult.}
\vspace{-0.3cm}
\label{fig1}
\end{figure}

To accelerate inference speed, a common approach is to reduce redundant parameters through various model compression techniques, such as pruning~\cite{han2015learning}, knowledge distillation~\cite{hinton2015distilling}, and model quantization~\cite{han2015deep}. 
Pruning, in particular, has attracted significant attention due to its ease of implementation on original models~\cite{filters2016pruning, fang2023depgraph}.
As represented in Fig.~\ref{fig1} (a), compared to general domain images, remote sensing images have a top-down perspective and variations in object scale~\cite{li2020object, maggiori2016convolutional, mahabir2018critical}, which makes feature extraction more challenging. As a result, the differences in channel importance become less pronounced~\cite{li2020object, zhang2020improved, zhang2019multi}. Therefore, although existing methods achieve remarkable compression performance in general domains~\cite{zhang2020pruning, wei2022based, hu2023data}, their performance in remote sensing tasks still has shortcomings.

Furthermore, some researchers adopt or develop lightweight models to enhance inference speed, such as literature~\cite{pham2023light} and RSCNet~\cite{chen2022rscnet}.
However, while effective in reducing computational costs, these models frequently struggle with limited capabilities in feature extraction and learning~\cite{li2022cloud, gulat2012remote, gao2009atmospheric, hosseini2024comprehensive}.
Therefore, the performance of such models may decline when faced with difficult samples in remote sensing domain. For example, as shown in Fig.~\ref{fig1} (b), the presence of blurred images caused by atmospheric noise is challenging for lightweight models to learn and classify. Prioritizing challenging samples is essential for maintaining the performance of lightweight models.

To this end, we introduce a novel model pruning method for remote sensing image classification, named \textbf{RemoteTrimmer}. It is composed of a Channel Attention Pruning (\textbf{CAP}) strategy and an Adaptive Mining Loss (\textbf{AML)}. Specifically, we design CAP module to enhance channel importance differentiation by mapping the original model’s features into a channel attention space, contributing to more precise pruning. 
Next, we introduce the AML function to emphasize difficult samples during the fine-tuning of pruned models, improving overall performance. To validate the effectiveness of our approach, we conduct experiments on two remote sensing classification datasets with two different models. Extensive experiments demonstrate that our approach effectively reduces the model's parameters, resulting in faster inference speed while preserving strong performance across two datasets.

Our main contributions are highlighted as follows:
\begin{itemize}
    \item We propose a novel pruning method by amplifying the differences in importance of model channels. \textbf{To the best of our knowledge, this is the first pruning method for remote sensing image classification models.} 

    \item We propose an adaptive mining loss function for the fine-tuning process of pruned lightweight models. It can adaptively learn difficult samples to reduce the accuracy loss caused by pruning. 

    \item Our approach achieves state-of-the-art (SoTA) performance on the EuroSAT and UCMerced\_LandUse datasets. The accuracy of the model after pruning can even be higher than before pruning.
\end{itemize}

\section{Method}
\begin{figure}[t] 
\centerline{\includegraphics[width=1\linewidth]{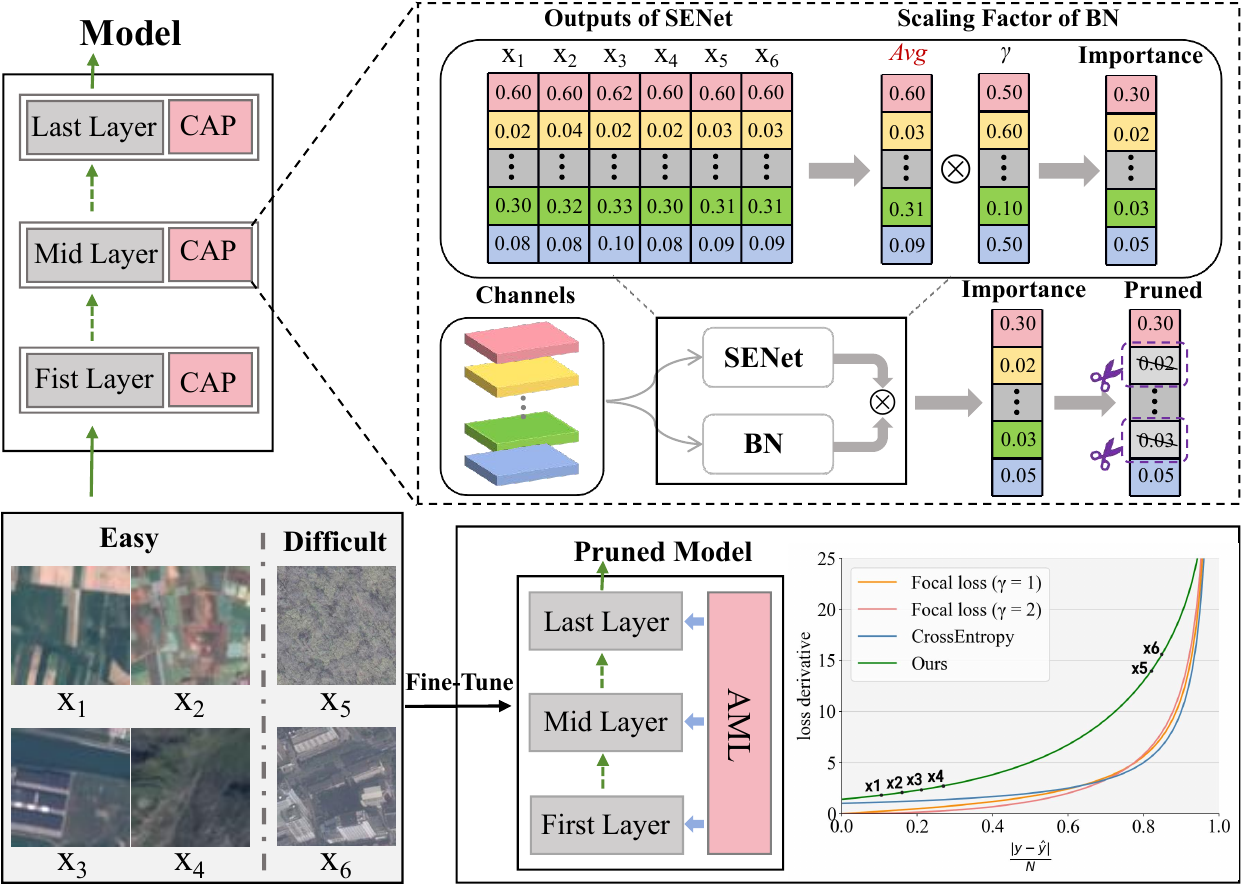}}
\caption{Overview of our RemoteTrimmer.  
In the pruning phase, we leverage intermediate outputs from SENet and scaling factors from the BN layer to map channel importance into the attention space. During the post-pruning fine-tuning phase, we design a lateral inhibition loss function to emphasize difficult samples. Our method effectively addresses two key challenges in remote sensing model pruning: the lack of distinct channel importance and the prevalence of difficult samples. 
}
\label{fig2}
\end{figure}

In this section, we introduce our proposed efficient structural pruning approach for remote sensing image classification, \textbf{RemoteTrimmer}. The overall framework is represented in~\ref{fig2}.

\subsection{Channel Attention Pruning} 
The significant scale variation of objects in remote sensing images, combined with the less distinct object features compared to general domains, results in reduced inter-channel differences within the model~\cite{li2020object, zhang2020improved}. The situation may adversely affect the performance of general pruning algorithms. Therefore, we propose to amplify the differences in channel importance by mapping the channel features to a new space.

Channel attention allows the model to dynamically adjust the importance of each channel~\cite{hassanin2024visual}. Inspired by it, we incorporate a channel attention module into the pruning process, mapping channel features into a new attention space. In this space, the contribution of each channel is amplified, which helps in identifying and pruning less significant channels.
Specifically, during the pruning phase, we leverage the intermediate output of  Squeeze-and-Excitation Network (SENet)~\cite{hu2018squeeze} to adjust the scaling factor $\gamma$ in the atch Normalization (BN) layer.
This adjustment enables $\gamma$ to focus on the more important channels.
Utilizing ResNet18~\cite{he2016deep} as an example, the process involves the following steps:

First, we apply the SENet module to ResNet18 and train the network. Then, we extract the intermediate output of SENet. The intermediate output at the $i$-th convolutional layer of SENet can be expressed as follows:
\begin{equation}
    s_i = \sigma \left( W_2 \cdot \text{ReLU} \left( W_1 \cdot \text{AvgPool}(X_i) \right) \right),
\end{equation}
where $X_i$ is the feature map output from the $i$-th convolutional layer, $\text{AvgPool}(X_i)$ represents the average pooling function, 
$W_1$ and $W_2$ are the weight matrices of the fully connected layers, ReLU denotes the ReLU activation function~\cite{glorot2011deep}, and $\sigma$ represents the sigmoid activation function.

\begin{table*}[t]
\centering
\caption{Comparison of pruning methods on EuroSAT and UCM datasets using ResNet18 and VGG16 models.}
\label{T1}
\setlength{\tabcolsep}{3.5mm}{
\begin{tabular}{c|c|ccc|ccc|c}
\toprule
\multicolumn{1}{c|}{\multirow{2}{*}{\textbf{Model}}} & 
\multicolumn{1}{c|}{\multirow{2}{*}{\textbf{Method}}}& 
\multicolumn{3}{c|}{\textbf{EuroSAT}} & 
\multicolumn{3}{c|}{\textbf{UCM}} & 
\multicolumn{1}{c}{\multirow{2}{*}{\textbf{Parameters}}} \\ \cline{3-8}
 &  & \textbf{Acc} & \textbf{$\pm \Delta$} & \textbf{MACs} & \textbf{Acc} & \textbf{$\pm \Delta$} & \textbf{MACs} & \textbf{} \\ \hline
\multirow{6}{*}{ResNet18} 
 & \cellcolor[HTML]{E0E0E0}baseline & \cellcolor[HTML]{E0E0E0}0.870 & \cellcolor[HTML]{E0E0E0}- & \cellcolor[HTML]{E0E0E0}0.15 G & \cellcolor[HTML]{E0E0E0}0.849 & \cellcolor[HTML]{E0E0E0}- & \cellcolor[HTML]{E0E0E0}2.38 G & \cellcolor[HTML]{E0E0E0}11.18 M \\ \cline{2-9}
 & BN~\cite{liu2017learning} & 0.870 & 0 & \multirow{5}{*}{0.02 G} & 0.830 & -0.019 & \multirow{5}{*}{0.24 G} & \multirow{5}{*}{1.00 M} \\ 
 & L1-norm~\cite{li2016pruning} & 0.872 & +0.002 & & 0.833 & -0.016 & & \\ 
 & FPGM~\cite{he2019filter} & 0.882 & +0.012 & & 0.838 & -0.011 & & \\ 
& DepGraph~\cite{fang2023depgraph} & 0.882 & +0.012 & & 0.830& -0.019 & & \\ 
 &\cellcolor[HTML]{DAE8FC}\textbf{Ours} & \cellcolor[HTML]{DAE8FC}\textbf{0.922} & \cellcolor[HTML]{DAE8FC}\textbf{+0.052} & & \cellcolor[HTML]{DAE8FC}\textbf{0.853} & \cellcolor[HTML]{DAE8FC}\textbf{+0.004} & & \\ \hline
\multirow{6}{*}{VGG16} 
 & \cellcolor[HTML]{E0E0E0}baseline & \cellcolor[HTML]{E0E0E0}0.957 & \cellcolor[HTML]{E0E0E0}- & \cellcolor[HTML]{E0E0E0}1.38 G & \cellcolor[HTML]{E0E0E0}0.903 & \cellcolor[HTML]{E0E0E0}- & \cellcolor[HTML]{E0E0E0}20.24 G & \cellcolor[HTML]{E0E0E0}134.31M\\ 
 \cline{2-9}
 & BN~\cite{liu2017learning} & 0.954 & -0.003 & \multirow{5}{*}{0.16 G} & 0.861 & -0.042 & \multirow{5}{*}{1.87 G} & \multirow{5}{*}{48.90 M} \\ 
 & L1-norm~\cite{li2016pruning} & 0.955 & -0.002 & & 0.848 & -0.055 & & \\ 
 & FPGM~\cite{he2019filter} & 0.955 & -0.002 & & 0.857 & -0.046 & & \\ 
 & DepGraph~\cite{fang2023depgraph} & 0.953 & -0.004 & & 0.862 & -0.041 & & \\  
 & \cellcolor[HTML]{DAE8FC}\textbf{Ours} & \cellcolor[HTML]{DAE8FC}\textbf{0.957} & \cellcolor[HTML]{DAE8FC}\textbf{0} & & \cellcolor[HTML]{DAE8FC}\textbf{0.872} & \cellcolor[HTML]{DAE8FC}\textbf{-0.031} & & \\ 
 \bottomrule
\end{tabular}
}
\end{table*}

Next, the channel attention scores are averaged across all selected samples. The vector $\bar{s} = [\bar{s}_1, \bar{s}_2, \dots, \bar{s}_C]$ captures the general importance of each channel across the dataset.

Finally, we combine these channel attention scores $\bar{s}_k$ with the scaling factor $\gamma_k$ from the Batch Normalization (BN) layer, which inherently reflects the relative importance of the channels in the network. The final channel importance score $I_k$ for $k$-th channel in $i$-th layer is computed as:
\begin{equation}
I_{i,k} = \bar{s}_{i,k} \cdot \gamma_{i,k}.
\end{equation}

Channels with higher $I_{i,k}$ values are considered more important during the pruning process, while channels with lower scores are pruned. For a pruning rate of $\alpha$ and a convolutional layer $C$, the pruning process can be expressed as: 
\begin{equation}
    \text{P} \left( C \right) = \{ c_i \in C \mid I(c_i) \leq \text{Q}_{\alpha}(I(C)) \},
\end{equation}
Where $c_i$ is the  $i$-th channel within $C$, and $I(c_i)$ denotes the importance score of channel $c_i$. The term $\text{Quantile}_{\alpha}(I(C))$ refers to the value below which the importance scores of $\alpha$ proportion of channels in $C$ fall.

At this stage, the model's original inter-layer relationships have been disrupted, leading to a significant drop in accuracy. Fine-tuning the model is necessary to restore its accuracy as closely as possible to the original level.

\subsection{Adaptive Mining Loss}
After pruning, the model's feature extraction capability significantly diminishes due to the reduced number of parameters, leading to prediction failures on difficult samples. To address this issue, we propose an Adaptive Mining Loss (AML) function for the fine-tuning process of the pruned model. 

Considering that the decline in feature extraction capability has a smaller impact on simpler samples, we can focus the fine-tuning process on learning difficult samples to compensate for the accuracy loss caused by pruning. In other words, we aim to give greater influence to the loss function from difficult samples that are predicted incorrectly.
The AML function can be expressed as:
\begin{equation}
    \mathcal{L}_{AM}(\mathit{y}, \hat{\mathit{y}}) = r \cdot \mathcal{L}_{CE}(y, \hat{y}) + e^{\frac{|y - \hat{y}|}{N}+ (\frac{|y - \hat{y}|}{N})^2},
\end{equation}
where $y$ is the true target value, $\hat{y}$ is the predicted value, $\theta$ represents the parameters of the classification model, $\mathcal{L}_{CE}$ is the cross-entropy loss, $r$ is weight for $\mathcal{L}_{CE}$ term in the loss function, and $N$ is the total number of classes in the classification dataset.

To provide a more intuitive understanding of the exponential function's impact and our loss function, we plot the loss curve of $\mathcal{L}_{LI}$ and its derivative with respect to a certain pixel of the model prediction $\hat{y}$ across different prediction errors $d$ in Fig. \ref{fig2}, comparing it with Cross-Entropy and Focal Loss~\cite{lin2017focal}.

Through the adaptive selection of difficult samples, we address the shortcomings in feature extraction capabilities of the pruned lightweight model during fine-tuning. Additionally, due to the changes in the target loss function, our method can unlock the potential of the pruned model, potentially exceeding the accuracy of the original model.


\section{EXPERIMENTS}

\subsection{Datasets and Evaluation Metrics}
To verify the effectiveness of our proposed approach, we adopted EuroSAT~\cite{helber2019eurosat} and UC Merced Land-Use (UCM)~\cite{qu2016deep} for experiments.
EuroSAT consists of 27,000 satellite images, covering 10 different land use classes. 
UCM is a high-resolution remote sensing image dataset, comprising 2,100 images across 21 categories. 
Follow the previous work~\cite{fang2023depgraph, wang2024towards}, we utilized the Accuracy (Acc), Multiply–Accumulate Operations (MACs) and Parameters as the evaluation metrics on model performance.

\begin{table}[b]
\centering
\caption{Effectiveness Analysis of CAP and AML on EuroSAT.} 
\label{T2}
\begin{tabular}{c|c|
>{\centering\arraybackslash}p{1.2cm}
>{\centering\arraybackslash}p{1.2cm}|
>{\centering\arraybackslash}p{1.2cm}}
\toprule
\textbf{Datasets} & \textbf{Model} & \textbf{CAP} & \textbf{AML} & \textbf{Acc} \\ 
\hline
\multirow{8}{*}{EuroSAT} 
 & \multirow{4}{*}{ResNet18} & & & 0.870\\  
 & & & $\checkmark$ & 0.872\\ 
 & & $\checkmark$ & & 0.917\\  
 & & \cellcolor[HTML]{DAE8FC}$\checkmark$ & \cellcolor[HTML]{DAE8FC}$\checkmark$ & \cellcolor[HTML]{DAE8FC}\textbf{0.921} \\ \cline{2-5} 
 & \multirow{4}{*}{VGG16} & & & 0.954\\ 
 & & & $\checkmark$ & 0.956\\ 
 & & $\checkmark$ & & 0.956\\  
 & & \cellcolor[HTML]{DAE8FC}$\checkmark$ & \cellcolor[HTML]{DAE8FC}$\checkmark$ & \cellcolor[HTML]{DAE8FC}\textbf{0.957}\\  \bottomrule
\end{tabular}
\end{table}

\subsection{Experimental Setup}


We adopted ResNet18~\cite{he2016deep} and VGG16~\cite{simonyan2014very} as the baseline models. Except for the experiments validating the pruning rate, all other pruning rates were set to 0.7. All experiments were conducted in PyTorch with an NVIDIA RTX 3090 GPU. 
During the fine-tuning phase, ResNet18 employed a batch size of 64 for 40 epochs, and VGG16 deployed a batch size of 16 for 20 epochs. All models were trained leveraging an SGD optimizer~\cite{cherry1998sgd}. The learning rate was initialized at 0.001 and adjusted employing an exponential decay method, with decay factors of 0.9 applied at 50\% and 75\% of the total epochs.

\subsection{Experimental Results and Analyses}
To explore the superiority of our pruning approach, we conducted comparative experiments with previous SoTA pruning methods on ResNet18 and VGG16. As shown in Tab. \ref{T1}, our method achieved SoTA performance on both the EuroSAT and UCM datasets. For example, on the EuroSAT dataset, ResNet18 achieved an accuracy that was 4.0\% higher than the previous SoTA method DepGraph, with a 5.2\% improvement over the pre-pruning model. On the UCM dataset, ResNet18 outperformed the SoTA method FPGM by 1.5\% and improved by 0.4\% compared to the pre-pruning model. 
The remarkable improvements reveal the effectiveness of our approach.

\subsection{Ablation Studies}
In this section, we validated the effectiveness of each module (CAP and AML), compared the performance of different loss functions, examined the impact of different attention mechanisms, conducted ablation on different pruning rates, and analyzed the influence of balance factors.

\subsubsection{Effectiveness of the Pruning and Fine-tuning Strategies}
We investigated the effectiveness of our proposed CAP and LIL components, as illustrated in Tab. \ref{T2}. By employing CAP, the accuracy improved by 4.7\% on ResNet18 and by 0.2\% on VGG16. The performance can be attributed to amplifying the differences in importance of model channels. After adopting AML, the accuracy further increased on two models. The experimental results demonstrate that both CAP and AML are effective and contribute to the final performance.

\begin{table}[t]
\centering
\caption{comparison of different loss functions on EuroSAT.}
\label{T3}                               
\renewcommand\arraystretch{1.2}
\begin{tabular}{cc|cc}
\toprule
\multicolumn{2}{c}{ \textbf{ResNet18}} & \multicolumn{2}{c}{\textbf{VGG16}} \\ \cline{1-4} 
\textbf{Loss} & \textbf{Acc} & \textbf{Loss} & \textbf{Acc} \\ \hline
CrossEntropy & 0.917 & CrossEntropy & 0.956 \\ 
Focal\_Loss~($\gamma$=1)~\cite{lin2017focal} & 0.914 & Focal\_Loss~($\gamma$=1)~\cite{lin2017focal} & 0.944 \\ 
Focal\_Loss~($\gamma$=2)~\cite{lin2017focal} & 0.904 & Focal\_Loss~($\gamma$=2)~\cite{lin2017focal} & 0.933 \\ 
\cellcolor[HTML]{DAE8FC}\textbf{Ours} & \cellcolor[HTML]{DAE8FC}\textbf{0.921} & \cellcolor[HTML]{DAE8FC}\textbf{Ours} & \cellcolor[HTML]{DAE8FC}\textbf{0.957} \\  
\bottomrule
\end{tabular}
\end{table}

\begin{table}[t]
\centering
\caption{Parameter Analysis of Balance Factor on EuroSAT.}
\label{T4}
\renewcommand\arraystretch{1.1}
\setlength{\tabcolsep}{3.5mm}{
\begin{tabular}{c|c|c|c|c|c}
\toprule
\textbf{Model} & \multicolumn{5}{c}{\textbf{Results}} \\
\hline
\multirow{2}{*}{ResNet18} & r & 0.1 & 0.4 & 0.7 & 1.0 \\ \cline{2-6}
 & Acc & 0.914 & \cellcolor[HTML]{DAE8FC}\textbf{0.922} & 0.918 & 0.915 \\ \hline
\multirow{2}{*}{VGG16} & r & 0.1 & 0.4 & 0.7 & 1.0 \\ \cline{2-6}
 & Acc & 0.951 & \cellcolor[HTML]{DAE8FC}\textbf{0.957} & 0.955 & 0.956 \\
\bottomrule
\end{tabular}
}
\end{table}

\subsubsection{Comparison of Different Loss Functions}
We compared our AML with various other similar loss functions like Focal loss, as represented in the Tab.~\ref{T3}. It can be observed that both models achieved better performance during the fine-tuning phase using our AML than with Cross Entropy loss and different parameter settings of Focal loss. This result may be due to our loss function's ability to better adaptively identify difficult samples and emphasize learning from them.

\subsubsection{Impact of Different Attention Mechanisms}
\begin{figure}[t] 
\centerline{\includegraphics[width=1\linewidth]{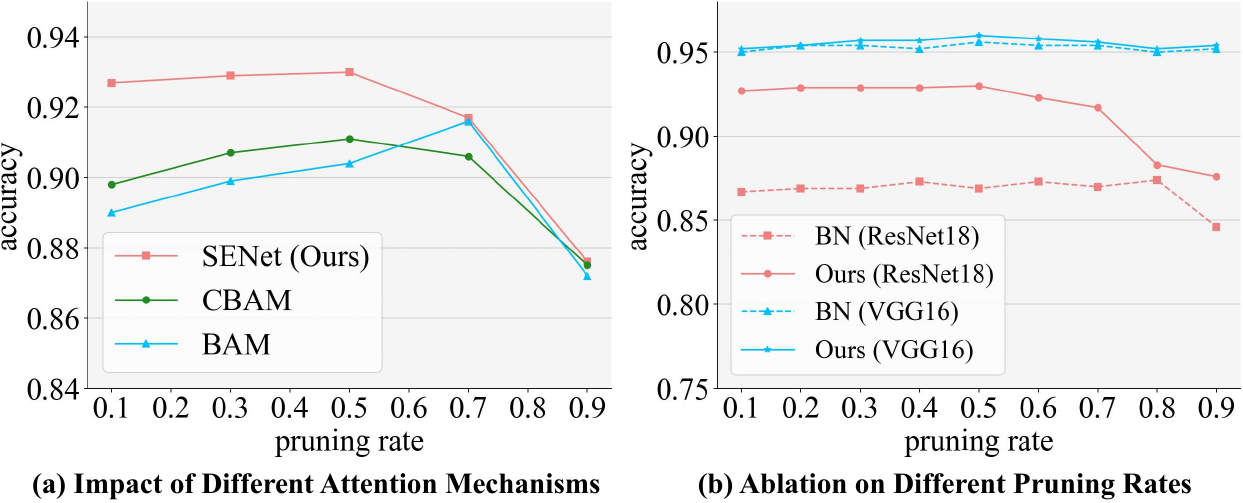}}
\caption{Ablation studies on different attention mechanisms (a) and pruning rates (b) on EuraSAT dataset. Our approach achieved the best accuracy across all pruning rates.
}
\vspace{-0.1cm}
\label{fig3}
\end{figure}

We investigated the influence of different attention spaces in our CAP module in Fig.~\ref{fig3} (a). It can be observed that SENet significantly outperformed other attention mechanisms at various pruning rates. The results indicate that channel attention is more conducive to amplifying the importance differences between channels. However, due to consideration of spatial characteristics, other types of attention mechanisms~\cite{woo2018cbam, park2018bam} are not conducive to channel discrimination.

\subsubsection{Ablation on Different Pruning Rates on EuroSAT}
We conducted experiments with varying pruning rates on the EuroSAT dataset utilizing ResNet18 and VGG16, as illustrated in Tab. \ref{T4}. It can be observed that our method demonstrates strong performance at various pruning rates for both models. Notably, there is a significant decline in accuracy for the ResNet18 model when the pruning rate reaches 0.9. This is due to the smaller scale of ResNet18, where a pruning rate of 0.9 results in too few model parameters, leading to a decrease in accuracy.


\subsubsection{Parameter Analysis of Balance Factors}
To ascertain the optimal balance factor value for $r$ in the adaptive mining loss function, we conducted parameter analysis on EuroSAT dataset with ResNet18 and VGG16 as the base model. The results are illustrated in Tab~\ref{T4}. When $r$ is set to 0.4, the model achieved the optimal accuracy of 92.2\%. Therefore, we ultimately adopted this set of values in our method.


\section{Conclusion}

In this paper, we proposed a high-precision model pruning algorithm, specifically designed for remote sensing image classification, which integrates channel attention mechanisms in the pruning phase with adaptive training in the fine-tuning phase. In the pruning stage, SENet architecture was utilized to map channel features into an attention space, which contributes to the precise differentiation of channel importance.
Regarding the fine-tuning stage, we introduced an adaptive mining loss function that guides models to focus on difficult samples. Experimental results demonstrated the integration of these two components ensures that the model maintains high accuracy even after significant pruning.

\clearpage
\bibliographystyle{IEEEtran}
\bibliography{IEEEexample}

\end{document}